\documentclass{amia}
\usepackage{lipsum} 
\usepackage{bm}
\usepackage{times}
\usepackage{helvet}
\usepackage{courier}
\usepackage{times}
\usepackage{latexsym}
\usepackage{amsmath}
\usepackage{algorithm}
\usepackage{algpseudocode}
\usepackage{url}
\usepackage{multirow}
\usepackage{enumitem}
\usepackage{array}
\usepackage[symbol]{footmisc}
\usepackage{threeparttable}
\usepackage{graphicx}
\usepackage{dblfloatfix} 
\usepackage{caption}
\usepackage[labelformat=simple]{subcaption}

\usepackage{adjustbox}
\usepackage{multirow}
\usepackage{hyperref}
\usepackage{xcolor}
\usepackage{float}
\usepackage{longtable}
\usepackage{amsmath}
\usepackage{amssymb}
\usepackage{subcaption}
\usepackage{tabularx}
\usepackage{makecell}
\usepackage{mathtools}

\newcolumntype{P}[1]{>{\centering\arraybackslash}p{#1}}

\begin{document}

\title{Automated Identification of Eviction Status from Electronic Health Record Notes}



\author{
Zonghai Yao, MSc$^1$, Jack Tsai, PhD, MSCP$^{2, 3, 4}$, Weisong Liu, PhD$^{5, 6, 7}$, David A. Levy, MD$^{6}$, Emily Druhl, MPH$^7$, Joel I Reisman, BA$^7$, Hong Yu, PhD$^{1, 5, 6, 7}$
}

\institutes{
    $^1$ Manning College of Information and Computer Sciences, University of Massachusetts Amherst, MA, USA\\
    $^2$ National Center on Homelessness among Veterans, U.S. Department of Veterans Affairs Homeless Programs Office, Washington, DC, USA\\
    $^3$ School of Public Health, University of Texas Health Science Center at Houston, Houston, TX, USA\\
    $^4$ Department of Psychiatry, Yale University School of Medicine, New Haven, CT, USA\\
    $^5$ Department of Medicine, University of Massachusetts Medical School, Worcester, MA, USA\\
    $^6$ Center for Biomedical and Health Research in Data Sciences, Miner School of Computer and Information Sciences, University of Massachusetts Lowell, MA, USA\\
    $^7$ Center for Healthcare Organization and Implementation Research, VA Bedford Health Care, MA, USA\\
}

\maketitle




\section*{Abstract}
\textbf{Objective:} Evictions are important social and behavioral determinants of health. Evictions are associated with a cascade of negative events that can lead to unemployment, housing insecurity/homelessness, long-term poverty, and mental health problems. In this study, we developed a natural language processing system to automatically detect eviction status from electronic health record (EHR) notes.  

\textbf{Materials and Methods:} 
We first defined eviction status (eviction presence and eviction period) and then annotated eviction status in 5000 EHR notes from the Veterans Health Administration (VHA). 
We developed a novel model, KIRESH,  that has shown to substantially outperform other state-of-the-art models such as fine-tuning pre-trained language models like BioBERT and Bio\_ClinicalBERT. 
Moreover, we designed a novel prompt to further improve the model performance by using the intrinsic connection between the two sub-tasks of eviction presence and period prediction. 
Finally, we used the Temperature Scaling-based Calibration on our KIRESH-Prompt method to avoid over-confidence issues arising from the imbalance dataset.

\textbf{Results:} 
KIRESH-Prompt substantially outperformed strong baseline models including fine-tuning the Bio\_ClinicalBERT model to achieve 0.74672 MCC, 0.71153 Macro-F1, and 0.83396 Micro-F1 in predicting eviction period and 0.66827 MCC, 0.62734 Macro-F1, and 0.7863 Micro-F1 in predicting eviction presence. 
We also conducted additional experiments on a benchmark social determinants of health (SBDH) dataset to demonstrate the generalizability of our methods.

\textbf{Conclusion and Future Work:} 
KIRESH-Prompt has substantially improved eviction status classification. 
We plan to deploy KIRESH-Prompt to the VHA EHRs as an eviction surveillance system to help address the US Veterans' housing insecurity. 

\textbf{Keywords:} Social and behavioral determinants of health, eviction, knowledge injection, prompt design, calibration


\section*{Introduction}
Evictions are important social and behavioral determinants of health (SBDH) that can negatively impact the health and lives of individuals and the communities in which they live \cite{desmond2016evicted, tsai2019systematic}. Evictions are associated with a cascade of negative events that can lead to unemployment, housing insecurity/homelessness, long-term poverty, and mental health problems \cite{tsai2019systematic, desmond2015eviction}. 
Since eviction can be considered a modifiable variable, proper interventions can help prevent housing insecurity/homelessness. However, unlike the widely-studied housing insecurity \cite{cutts2011us} and homelessness \cite{hwang2001homelessness}, we found little research on eviction.

Eviction has been difficult to study because there currently exists no national, centralized database of evictions. In addition, eviction is a process with varying timelines so studies using cross-sectional data face limitations in fully capturing eviction cases.
Moreover, there has been almost no study of eviction using large electronic health record (EHR) systems. A major impediment to such research is that most EHR systems do not contain structured codes to capture evictions or eviction risk. The International Classification of Diseases, Tenth Revision (ICD-10) contains a code for homelessness (Z59.0), but there is no code for eviction per se.

Our project aims to develop natural language processing (NLP) models that can automatically detect eviction status in Veterans’ EHR notes. We first annotated 5000 EHR notes from the Veterans Health Administration (VHA) and identified a total of 5,271 mentions of eviction status. We then proposed two multi-class learning tasks: eviction presence and period predictions. The eviction presence prediction contains 6 classes: Absent, Present, Pending, Mutual Rescission (MR), Uncertain, and Irrelevant. The eviction period prediction contains 5 classes: Current, History, Future/Hypothetical, Uncertain, and No. Both tasks are challenging because the data are heavily imbalanced. As shown in Table \ref{table:data-distribution}, while class ``Present’' and ``Current’' have the highest incidences of 2503 and 2734, class ``Mutual Rescission’' and ``Uncertain’' have only 91 and 62. 

We first tried two baseline models that have been shown to be state-of-the-art for biomedical text classification tasks. One is to use the pre-trained text representation BioSentVec \cite{chen2019biosentvec} as the sentence embeddings and the logistic regression \cite{cox1958regression} as the classifier. The other is to directly fine-tune pre-trained language models (BioBERT \cite{lee2020biobert} and Bio\_ClinicalBERT \cite{alsentzer2019publicly}) on our dataset.

A novel contribution of our work is that we leverage other adverse SBDH to help improve eviction detection. Eviction is associated with other SBDH \cite{janelle2018social}. For example, eviction leads to housing insecurity and homelessness. Financial insecurity, which may be caused by unemployment and mental health problems, may lead to eviction. 
\cite{Mitra2022SBDHandSuicide} found that patients frequently have multiple SBDH, and collectively they have a stronger association with an increased risk of death by suicide than the association with single SBDH. 
No research has examined how SBDH and multiple SBDH can be used for the prediction of another SBDH. Since the main topic of this paper is to identify eviction, we developed a novel method, called Knowledge Injection based on Ripple Effects of Social and Behavioral Determinants of Health (KIRESH), which leverages the associations between eviction and other SBDH. 
Furthermore, we designed a special prompt using the intrinsic connection between the two tasks of eviction presence and period prediction to help the model self-reason during the training.

Our experiments show that our proposed methods consistently outperformed the baseline methods.
In order to mitigate over-confidence issues arising from the imbalance dataset, we also added the Temperature Scaling method for model calibration, where Expected Calibration Error (ECE) can be used to measure the confidence of the model predictions.

Our contributions are as follows:

\begin{enumerate}[topsep=0.5pt,itemsep=0.2ex,partopsep=0.2ex,parsep=.20ex, label = $\ast$]
  \item We defined eviction status and annotated the status using an EHR note dataset.
  \item We introduced a novel Knowledge Injection based on Ripple Effects of Social and Behavioral Determinants of Health (KIRESH) method, which improved eviction status detection using co-occurrent SBDH.
  \item We designed a novel KIRESH with prompt to further improve the eviction status prediction by leveraging 
  the intrinsic connection 
  between eviction status (presence and period).
  \item We validated the generalizability of KIRESH with other SBDH on an external dataset.
  \item Finally, with the method of Temperature Scaling, KIRESH was calibratable and predicted with confidence.
\end{enumerate}

\section*{Related Work}
\subsubsection*{Social and Behavioral Determinants of Health Detection}
SBDH identification has been an active research field. \cite{uzuner2008identifying} detected smoking in the i2b2 Challenge, and \cite{alzoubi2018automated} detected alcohol consumption. 
Many approaches \cite{wicentowski2008using, carrero2006quick, pedersen2006determining, jonnagaddala2015preliminary} used pre-trained textual representations \cite{topaz2019extracting} such as word2vec \cite{mikolov2013efficient}, phrase2vec \cite{wu2020phrase2vec}, and BioSentVec \cite{chen2019biosentvec} as features and logistic regression \cite{cox1958regression} as a classifier for these classification tasks. 
Some work has also been done in detecting homelessness using EHRs \cite{gundlapalli2013using}. 
Other studies predicted the status of multiple SBDH statuses in EHR notes \cite{Mitra2022SBDHandSuicide, ahsan2021mimic}. 

\subsubsection*{Eviction and House Insecurity}
Evictions are important SBDH\cite{vasquez2017threat}, yet to our knowledge, there has been little work on the detection of eviction status. In contrast, most works developed NLP models to detect housing insecurity\cite{Mitra2022SBDHandSuicide} and homelessness\cite{gundlapalli2013using, gundlapalli2014extracting}.
Eviction, housing insecurity, and homelessness are intertwined \cite{fetzer2019housing}. 
Eviction leads to housing insecurity/homelessness, but not always \cite{treglia2023quantifying}. 
They can be considered different constructs. 
Having both eviction and housing insecurity constructs provides a comprehensive system of monitoring and capturing cases at risk for housing insecurity/homelessness, including evictions, among other common health factors (e.g., mental illness, substance abuse, etc.). 
So in this paper, we don't consider eviction as a subtype of housing insecurity/homelessness but as a new SBDH. 

\subsubsection*{Text Classification using NLP and BioNLP}
In the NLP and BioNLP domains, the standard methods of classification tasks have recently been shifted to the pre-train and fine-tune paradigm \cite{kotsiantis2007supervised, radford2018improving, peters-etal-2018-deep}.
In this paradigm, language models can be trained on large datasets. In this process language models learn robust language-related features, which can be fine-tuned for the downstream applications \cite{devlin2018bert, kwon2022medjex, yao2020zero}.
In this paper, we followed the pre-train and fine-tune paradigm to train our KIRESH model. We compared it with state-of-the-art baseline models.
In addition, it is well known that modern neural networks including language models are poorly calibrated, especially in the extremely imbalanced setting \cite{guo2017calibration, jiang2021can, frenkel2021network}. They tend to overestimate or underestimate probabilities when compared to the expected accuracy. This results in misleading reliability and therefore corrupts the decision policy. Temperature scaling is a post-processing method that fixes this \cite{guo2017calibration}\footnote{More related work can be found in the \nameref{appendix:related-work}}. Our results have shown that this simple yet effective method did not affect the model’s accuracy, yet avoided the over-confidence issues arising from the imbalanced dataset. 

\begin{table}[!htbp]
\begin{tabularx}{\textwidth}{|p{3.5cm}|X|}

\hline
History/absent & 
\begin{enumerate}[topsep=0.5pt,itemsep=0.2ex,partopsep=0.2ex,parsep=.20ex, label = $\ast$]
    \item "background check shows Veteran was never evicted"
\end{enumerate}
\\

\hline

History/present & 
\begin{enumerate}[topsep=0.5pt,itemsep=0.2ex,partopsep=0.2ex,parsep=.20ex, label = $\ast$]
    \item "has been evicted"
    \item "was evicted from previous residence"
\end{enumerate}
\\

\hline
History/uncertain & 
\begin{enumerate}[topsep=0.5pt,itemsep=0.2ex,partopsep=0.2ex,parsep=.20ex, label = $\ast$]
    \item "may have had an eviction"
\end{enumerate}
\\

\hline

Current/absent & 
\begin{enumerate}[topsep=0.5pt,itemsep=0.2ex,partopsep=0.2ex,parsep=.20ex, label = $\ast$]
  \item "landlord denies they are being evicted"
  \item "agreed not to evict him"
\end{enumerate}
\\
\hline

Current/present & 
\begin{enumerate}[topsep=0.5pt,itemsep=0.2ex,partopsep=0.2ex,parsep=.20ex, label = $\ast$]
    \item "is getting/being evicted"
    \item "received eviction notice and must be out by 5th February"
\end{enumerate}
\\
\hline

Current/pending
&
\begin{enumerate}[topsep=0.5pt,itemsep=0.2ex,partopsep=0.2ex,parsep=.20ex, label = $\ast$]
    \item "planning to evict"
    \item "trying to evict him"
    \item "intend to evict"
    \item "landlord has an eviction notice to serve her"
\end{enumerate}
\\
\hline

Current/uncertain
&
\begin{enumerate}[topsep=0.5pt,itemsep=0.2ex,partopsep=0.2ex,parsep=.20ex, label = $\ast$]
    \item "This SW phoned patient re: eviction"
    \item "if patient had been issued an eviction notice"
    \item "pt did not show up to eviction hearing" – eviction hearings can go either way
    \item "f/u regarding possible eviction"
\end{enumerate}
\\
\hline

Future/hypothetical 
&
\begin{enumerate}[topsep=0.5pt,itemsep=0.2ex,partopsep=0.2ex,parsep=.20ex, label = $\ast$]
    \item "requested paperwork for eviction prevention support"
    \item "afraid of an eviction"
    \item "can’t evict her during the pandemic"
\end{enumerate}
\\
\hline

Irrelevant &
\begin{enumerate}[topsep=0.5pt,itemsep=0.2ex,partopsep=0.2ex,parsep=.20ex, label = $\ast$]
    \item "patient has to evict his son"
    \item "Veteran’s girlfriend is getting evicted"
\end{enumerate}
\\
\hline

MR /present &
\begin{enumerate}[topsep=0.5pt,itemsep=0.2ex,partopsep=0.2ex,parsep=.20ex, label = $\ast$]
    \item "Landlord and Veteran agreed on mutual recission at end of month"
\end{enumerate}
\\
\hline

MR /pending &
\begin{enumerate}[topsep=0.5pt,itemsep=0.2ex,partopsep=0.2ex,parsep=.20ex, label = $\ast$]
    \item "Landlord states will talk with Veteran about mutual recission instead"
\end{enumerate}
\\
\hline

\end{tabularx}

\caption{Annotation Examples}
\label{table:annotation-examples}
        
\end{table}

\begin{table*}
    \centering
    \begin{tabular}{c|ccccc|cccccc}
    \hline
    
    & \multicolumn{5}{c|}{eviction period} & \multicolumn{6}{c}{eviction presence} \\
    
    \hline
    & future & irrelevant & current & uncertain & history & no & present & absent & uncertain & pending & MR
    \\
    \hline
    
    $all$ & 1188 & 406 & 2734 & 62 & 881 & 1656 & 2503 & 215 & 115 & 691 & 91
    \\
    \hline
    
    train & 718 & 227 & 1646 & 35 & 537 & 980 & 1504 & 130 & 66 & 428 & 55
    \\
    
    eval & 242 & 91 & 532 & 13 & 176 & 346 & 498 & 42 & 28 & 119 & 21
    \\
    
    test & 228 & 88 & 556 & 14 & 168 & 330 & 501 & 43 & 21 & 144 & 15
    \\
    
    \hline

    len & 42 & 48 & 37 & 27 & 34 & 43 & 34 & 43 & 35 & 40 & 47
    \\
    
    \hline
    
    \end{tabular}
    \caption{Data distribution for eviction status dataset. "len" represents word-level average length for different categories.}
    \label{table:data-distribution}
\end{table*}

\section*{Eviction status Dataset}
\subsection*{Data Collection Process}
We randomly selected 5,000  EHR notes (we focused on the following three types of notes: (Homeless Program Notes, Social Work Notes, and Mental Health Notes) from 2016 to 2021 in the Corporate data warehouse (CDW) \footnote{https://www.hsrd.research.va.gov/for\_researchers/cdw.cfm} database if the notes contained any of the following eviction initiation–related keywords, including ``notice to mutual rescission’', ``notice to pay rent’', ``notice to vacate’', ``writ of possession’', ``summary process’', ``summary judgment’', and ``unlawful detainer’'. 

\subsection*{Eviction Status Taxonomy}
Through an iterative process, two annotators (DL and ED) worked with the domain expert (JT) to create an eviction status taxonomy and finally annotated 5,271 mentions of eviction status in the sampled EHR notes. 
Table \ref{table:annotation-examples} shows the taxonomy and some examples of our annotation.
In the annotation, the unit of annotation was ``span’' but not ``sentence’'. Therefore, a sentence could have multiple labels because we chose multiple different spans within the sentence.
Finally, the results showed that there was good agreement among annotators (Fleiss’ kappa = 0.929), indicating high annotation reliability using this taxonomy. 
\footnote{More annotation examples and data agreement details can be found in the \nameref{appendix:anno-example}}.

\subsection*{Data Construction}
We grouped Eviction Presence into 6 categories: Absent, Present, Pending, Mutual Rescission (MR), Uncertain, and Irrelevant. We grouped Eviction Period into 5 categories: Current, History, Future/Hypothetical, Uncertain, and No. Note that some annotations only had eviction presence -- we assign all such data with "No" in the eviction period. The total annotated 5,271 eviction statuses were randomly split into training, validation, and test sets of 3163, 1054, and 1054, respectively. Table \ref{table:data-distribution} shows the data distribution.

\section*{Methods}
Eviction status classification can be formulated as multi-task learning. Specifically, considering an input sentence $s$ from an EHR note, the task is assigning labels $Y = [c_1, c_2]$ for eviction presence and period. Here, $c_1$ is for eviction presence, and $c_2$ is for eviction period. 

\subsection*{Baseline Models}
Our first baseline model used the most recent biomedical pre-trained textual representations, \textbf{BioSentVec} \cite{chen2019biosentvec}, created using PubMed and MIMIC-III, as our text representation. After the hidden representation from BioSentVec, we used \textbf{logistic regression} \cite{cox1958regression} to perform classification based on sentence embeddings.

Our second baseline model is fine-tuning an entire pre-trained language model (PLM). We first encode the input sentence $s$ into hidden representation $h$ with a PLM. 
In this paper, we use \textbf{BioBERT} \cite{lee2020biobert} and \textbf{Bio\_ClinicalBERT} \cite{alsentzer2019publicly} for all our experiments. 
BioBERT is initialized with BERT and trained on PubMed abstracts and PubMed Central full-text articles (PMC), 
and Bio\_ClinicalBERT is initialized with BioBERT and trained on the MIMIC-III notes.
After we get the representation $h$, two simple softmax classifiers are added to the top of PLM to predict the probability of label $c_1$ and $c_2$ respectively:

$$
p(c_1|h) = softmax(W_1 h)
$$
$$
p(c_2|h) = softmax(W_2 h)
$$

where $W_1$ and $W_2$ are the task-specific parameter matrix. We fine-tuned all the parameters from PLM as well as $W_1$ and $W_2$ jointly by maximizing the log-probability of the correct label.

\subsection*{Knowledge Injection based on Ripple Effects of Social and Behavioral Determinants of Health (KIRESH)}
As described previously, SBDH are not isolated determinants; 
they are intertwined. For example, a person who encounters job insecurity, house insecurity, food insecurity, financial insecurity, or other SBDH problems is often more likely to encounter an eviction problem. This is known as Ripple Effects of SBDH \cite{janelle2018social}. 
Our proposed KIRESH takes advantage of the intrinsic connection between eviction and other SBDH by injecting other SBDH into the model prediction.
Such patient-level SBDH are difficult for human experts to generate, therefore we adopted an AI-in-the-loop approach. Specifically,
We propose to identify SBDH by using a most recently developed NLP system \cite{Mitra2022SBDHandSuicide} that, 
given an input sentence $s$, outputs categories of 13 SBDH.
We then incorporated the identified ancillary SBDH information 
to the end of input $x_s$, and use a special token [SEP] to divide different SBDH. So the knowledge injection based input to the model will be:

\begin{center} $x_s=$ [CLS] s [SEP] [SBDH] \end{center}

where [SBDH] will be defined as ”SBDH1 [SEP] SBDH2 [SEP] ...”, which are the predictions generated by the SBDH prediction model.

\subsection*{Prompt Design for Multi-task Learning}
We design the prompt as a carrier to pass information between the two tasks to help the model train more stably.
First, we design the free text prompt template "The eviction presence is [MASK] . The eviction period is [MASK] ." for the eviction classification. 
Then we add the prompt to the end of the previous input and separate it from the previous content with [SEP]:

\begin{center} $x_s=$[CLS] s [SEP] [SBDH] [SEP] The eviction presence is [MASK] . The eviction period is [MASK] . \end{center}

In the experiment, we found that if any one of the eviction statuses is known, the prediction accuracy of the other will be greatly improved. 
This finding motivates us to design and use the above prompt and replace the [MASK] token of the eviction presence or eviction period with the corresponding label according to a certain probability during training.
Specifically, for a certain data in the training process, we ensure that the following three situations have the same possibility to be used as the input of the model \footnote{The probability here is a hyper-parameter, but we found that there is not much difference in the results of different probabilities. For convenience, we use 1:1:1.}:

\begin{enumerate}
    \item "[CLS] s [SEP] [SBDH] [SEP] The eviction presence is [MASK] . The eviction period is [MASK] ."
    \item "[CLS] s [SEP] [SBDH] [SEP] The eviction presence is [label1] . The eviction period is [MASK] ."
    \item "[CLS] s [SEP] [SBDH] [SEP] The eviction presence is [MASK] . The eviction period is [label2] ."
\end{enumerate}

In the evaluation and testing stage, the data will always be "[CLS] s [SEP] [SBDH] [SEP] The eviction presence is [MASK] . The eviction period is [MASK] ." 

\subsection*{Model Calibration by Temperature Scaling}
Modern neural networks are easily miscalibrated, especially in imbalanced datasets. In this paper, we use Temperature scaling to fix it \cite{guo2017calibration}. Temperature scaling introduces a positive scala temperature $T$ in
the final classification layer to make the probability distribution either more peaky or smooth, where it divides the logits by a learned scalar parameter.

$$ softmax = \frac{e^\frac{z}{T}}{\sum_{i=1}^{K} e^\frac{z_i}{T}} $$

where $z$ is the logit, which is the input to the softmax function. $K$ is the class number. $T$ is the learned parameter. We learn this parameter on a validation set and choose $T$ by minimizing the negative log likelihood loss. If $T$ is close to 0, the probability distribution becomes concentrated, while as $T$ approaches $\infty$, the probability distribution becomes uniform. Because the parameter $T$ does not change the maximum of the softmax function, temperature scaling does not affect the model’s accuracy. We hope to use this simple yet effective method to avoid large miscalibration while using various methods above.

    \begin{table*}
        \centering
        \begin{tabular}{c|ccc|ccc}
        \hline
        
        & \multicolumn{3}{c|}{eviction period} & \multicolumn{3}{c}{eviction presence} \\
        
        & MCC & Macro-F1 & Micro-F1 & MCC & Macro-F1 & Micro-F1
        \\
        \hline
        
        & \multicolumn{6}{c}{evaluation dataset} \\
        \hline
        
        & \multicolumn{6}{c}{100\% training data} \\
        \hline
        
        Logistic Regression & 0.42308 & 0.49737 & 0.59492 & 0.31322 & 0.35782 & 0.53391
        \\
        
        \hline
        
        BioBERT & 0.71112 & 0.68764 & 0.81326 & 0.62862 & 0.56955 & 0.75852
        \\
        
        BioBERT-KIRESH & 0.71886 & 0.68958 & 0.81758 & 0.63565 & 0.57614 & 0.76228
        \\
        
        BioBERT-Prompt & 0.71688 & \textcolor{red}{0.70933} & 0.81592 & 0.62890 & 0.57376 & 0.75900
        \\
        
        BioBERT-KIRESH-Prompt & \textcolor{red}{0.73157} & 0.68931 & \textcolor{red}{0.82527} & \textcolor{red}{0.63964} & \textcolor{red}{0.58380} & \textcolor{red}{0.76597}
        \\
        
        \hline
        
        Bio\_ClinicalBERT & 0.72168 & 0.70987 & 0.81851 & 0.64501 & 0.58458 & 0.76824
        
        \\
        
        Bio\_ClinicalBERT-KIRESH & 0.73169 & 0.70322 & 0.82571 & 0.65882 & 0.60815 & 0.77610
        \\
        
        Bio\_ClinicalBERT-Prompt & 0.73497 & 0.69306 & 0.82860 & 0.66017 & 0.61209 & 0.77736
        \\
        
        Bio\_ClinicalBERT-KIRESH-Prompt & \textcolor{red}{0.74672} & \textcolor{red}{0.71153} & \textcolor{red}{0.83396} & \textcolor{red}{0.66827} & \textcolor{red}{0.62734} & \textcolor{red}{0.78636}
        \\
        
        \hline
        
        & \multicolumn{6}{c}{10\% training data} \\
        \hline
        
        Bio\_ClinicalBERT & \textcolor{red}{0.48575} & \textcolor{red}{0.52754} & \textcolor{red}{0.66969} & \textcolor{red}{0.42374} & \textcolor{red}{0.33865} & \textcolor{red}{0.61986}
        \\
        
        Bio\_ClinicalBERT-KIRESH & 0.37397 & 0.45372 & 0.59829 & 0.37551 & 0.27973 & 0.59001
        \\
        
        Bio\_ClinicalBERT-Prompt & 0.45141 & 0.48631 & 0.64337 & 0.37235 & 0.30360 & 0.58824
        \\
        
        Bio\_ClinicalBERT-KIRESH-Prompt & 0.45836 & 0.51658 & 0.65262 & 0.39479 & 0.32333 & 0.60071
        \\
        
        \hline
        
        & \multicolumn{6}{c}{20\% training data} \\
        \hline
        
        Bio\_ClinicalBERT & \textcolor{red}{0.63535} & \textcolor{red}{0.62299} & \textcolor{red}{0.76107} & \textcolor{red}{0.55297} & 0.40236 & \textcolor{red}{0.70495}
        \\
        
        Bio\_ClinicalBERT-KIRESH & 0.62747 & 0.61804 & 0.75791 & 0.54645 & 0.4008 & 0.69571
        \\
        
        Bio\_ClinicalBERT-Prompt & 0.60741 & 0.60835 & 0.74300 & 0.53278 & \textcolor{red}{0.40441} & 0.69278
        \\
        
        Bio\_ClinicalBERT-KIRESH-Prompt & 0.61762 & 0.59745 & 0.75047 & 0.54101 & 0.38659 & 0.69334
        \\
        
        \hline
        
        & \multicolumn{6}{c}{30\% training data (about 1000 data)} \\
        \hline
        
        Bio\_ClinicalBERT & 0.61659 & \textcolor{red}{0.63114} & 0.75060 & 0.54723 & 0.42648 & 0.69227
        \\
        
        Bio\_ClinicalBERT-KIRESH & 0.62684 & 0.60366 & 0.75624 & 0.54037 & 0.44701 & 0.70389
        \\
        
        Bio\_ClinicalBERT-Prompt & 0.60997 & 0.61172 & 0.74684 & 0.54757 & 0.43659 & 0.69617
        \\
        
        Bio\_ClinicalBERT-KIRESH-Prompt & \textcolor{red}{0.63225} & 0.62746 & \textcolor{red}{0.76147} & \textcolor{red}{0.56838} & \textcolor{red}{0.49451} & \textcolor{red}{0.71654}
        \\
        
        \hline
        
        & \multicolumn{6}{c}{40\% training data} \\
        \hline
        
        Bio\_ClinicalBERT & 0.63513 & 0.63995 & 0.76566 & 0.56018 & 0.49380 & 0.71272
        \\
        
        Bio\_ClinicalBERT-KIRESH & 0.65049 & \textcolor{red}{0.6649} & 0.77226 & 0.57233 & 0.44642 & 0.71302
        \\
        
        Bio\_ClinicalBERT-Prompt & 0.63218 & 0.62825 & 0.75732 & \textcolor{red}{0.58804} & 0.46227 & 0.71286
        \\
        
        Bio\_ClinicalBERT-KIRESH-Prompt & \textcolor{red}{0.65376} & 0.63480 & \textcolor{red}{0.77510} & 0.57803 & \textcolor{red}{0.49697} & \textcolor{red}{0.72261}
        \\
        
        \hline
        
        & \multicolumn{6}{c}{50\% training data} \\
        \hline
        
        Bio\_ClinicalBERT & 0.67900 & 0.67326 & 0.79335 & 0.59531 & 0.51671 & 0.73812
        \\
        
        Bio\_ClinicalBERT-KIRESH & 0.66598 & 0.65271 & 0.7839 & 0.59673 & 0.52986 & 0.73524
        \\
        
        Bio\_ClinicalBERT-Prompt & 0.68332 & 0.66724 & 0.79380 & 0.59646 & 0.51494 & 0.73177
        \\
        
        Bio\_ClinicalBERT-KIRESH-Prompt & \textcolor{red}{0.68891} & \textcolor{red}{0.67656} & \textcolor{red}{0.80006} & \textcolor{red}{0.61120} & \textcolor{red}{0.53140} & \textcolor{red}{0.73996}
        \\
        
        \hline
        
        & \multicolumn{6}{c}{test dataset (100\% training data)} \\
        \hline
        
        Bio\_ClinicalBERT & 0.67969 & 0.67788 & 0.79083 & 0.60494 & 0.52284 & 0.74029
        \\
        
        Bio\_ClinicalBERT-KIRESH & 0.69234 & 0.65798 & 0.79814 & 0.60655 & 0.55068 & 0.74049
        \\
        
        Bio\_ClinicalBERT-Prompt & 0.70900 & \textcolor{red}{0.69209} & 0.81086 & 0.61809 & 0.53212 & 0.75071
        \\
        
        Bio\_ClinicalBERT-KIRESH-Prompt & \textcolor{red}{0.71903} & 0.68639 & \textcolor{red}{0.81607} & \textcolor{red}{0.6285} & \textcolor{red}{0.55136} & \textcolor{red}{0.75162}
        \\
        \hline
        
        \end{tabular}
        \caption{Experimental results of three sets of experiments: 1. Using all training data, we report the results of baseline models and several methods we proposed in the evaluation dataset; 2. Exploring Bio\_ClinicalBERT and several methods we proposed in different training data sizes; 3. the performance of Bio\_ClinicalBERT (the better PLM in evaluation dataset) with several methods we proposed in the test dataset.
        Red means the best performance.
        We provide confidence interval and prediction intervals of the models in Table \ref{Table:95ci-score-for-main-results}.}
        \label{Table:main-results}
    \end{table*}

    \begin{table*}
        \centering
        \begin{tabular}{c|cc|cc}
        \hline
        
        & \multicolumn{2}{c|}{eviction period} & \multicolumn{2}{c}{eviction presence} \\
        
        \hline
        
        Logistic Regression & \multicolumn{2}{c|}{4.703} & \multicolumn{2}{c}{2.611}
        \\
        \hline
        
        & No TS & TS & No TS & TS
        
        \\
        \hline
        
        BioBERT & 15.486 & 5.744 & 16.751 & 3.011
        \\
        
        BioBERT-KIRESH & 14.269 & 4.273 & 15.880 & 3.305
        \\
        
        BioBERT-Prompt & 16.351 & 7.619 & 19.197 & 7.127
        \\
        
        BioBERT-KIRESH-Prompt & 15.559 & 7.664 & 19.097 & 7.763
        \\
        \hline
        
        Bio\_ClinicalBERT & 16.638 & 5.123 & 16.604 & 3.023
        \\
        
        Bio\_ClinicalBERT-KIRESH & 14.655 & 4.300 & 14.878 & 3.318
        \\
        
        Bio\_ClinicalBERT-Prompt & 16.918 & 6.448 & 19.262 & 4.440
        \\
        
        Bio\_ClinicalBERT-KIRESH-Prompt & 15.240 & 5.433 & 17.693 & 6.739
        \\
        
        \hline
        
        \end{tabular}
        \caption{ECE (in \%) computed for without temperature scaling (No TS) and temperature scaling (TS).}
        \label{Table:Calibration}
    \end{table*}

\section*{Experiments}
We introduce the setting of the experiment in the \nameref{appendix:setting}

\subsection*{Metrics}
Both eviction presence and eviction period tasks are multi-class learning. For model prediction and label pairs, we calculate the confusion matrix C for both tasks, which are True Positives (TP), True Negatives (TN), False Positives (FP), and False Negatives (FN) for each class.
Both tasks are imbalanced, so we use two widely-used metrics, \textbf{Macro-F1} and \textbf{Micro-F1} for evaluation \cite{gorodkin2004comparing}. 
In addition, due to data imbalance, we also report the Matthews Correlation Coefficient \cite{gorodkin2004comparing} (\textbf{MCC}) score \footnote{https://scikit-learn.org/stable/modules/model\_evaluation.html\#matthews-corrcoef}, a more reliable score which produces a high score only if the prediction obtained good results in all of the four confusion matrix categories (TP, TN, FP, and FN). Finally, we report a scalar summary statistic of calibration. Expected Calibration Error \cite{guo2017calibration,naeini2015obtaining} – or \textbf{ECE} – partitions predictions into M equally-spaced bins and takes a weighted average of the bins' accuracy/confidence difference \footnote{More detailed information about our metrics is in \nameref{appendix:metrics}}.

\subsection*{Experimental Results}

\subparagraph{Baseline Models and KIRESH-Prompt} 
According to Table \ref{Table:main-results}, the results of fine-tuning on PLMs (BioBERT and Bio\_ClinicalBERT) greatly exceed the results of BioSentV and logistic regression. In Table \ref{Table:main-results}, we also show the effectiveness of KIRESH and our new prompt specially designed for this multi-task learning. Bio\_ClinicalBERT-KIRESH-Prompt achieves the highest scores on all metrics for both tasks. We will analyze the advantages of our proposed methods in detail in the Discussion section.

\subparagraph{Model’s Performance in Different Data Sizes}
Table \ref{Table:main-results} reports the model performance using different amounts of training data. The results show that when the amount of training data is small (10\% and 20\% of the total available training data), our proposed methods under-performed the Bio\_ClinicalBERT's fine-tuning model. 
When the training data is 30\% or higher, KIRESH-Prompt shows improvement across most of the evaluation metrics. 

\subparagraph{Model Calibration by Temperature Scaling}
Model Calibration by Temperature Scaling (TS) helps improve model calibration due to the data imbalance challenge. 
We report in Table \ref{Table:Calibration} the ECE of BioBERT and Bio\_ClinicalBERT with our methods before and after the Temperature Scaling-based Calibration. 
Our results show that pre-trained text representation and logistic regression have relatively low ECE with and without TS. The results demonstrate that the output confidence values are close to the actual accuracies, which means the logistic regression is well-calibrated even though its performance is sub-optimal.
Our results also show that the two PLMs have larger values of ECE, which indicates a larger difference between output confidence and actual model accuracy of the prediction.
The results show that adding other SBDH information as additional knowledge will reduce the ECE values, but adding our designed prompt has made the models substantially over-confident.
Fortunately, Temperature Scaling-based Calibration substantially reduces most models' ECE to a more reasonable value.

\section*{Discussion}

\subsection*{Ripple Effects of SBDH for Eviction Status}

    \begin{figure*}[htbp]
        \centering
        \includegraphics[width=\linewidth]{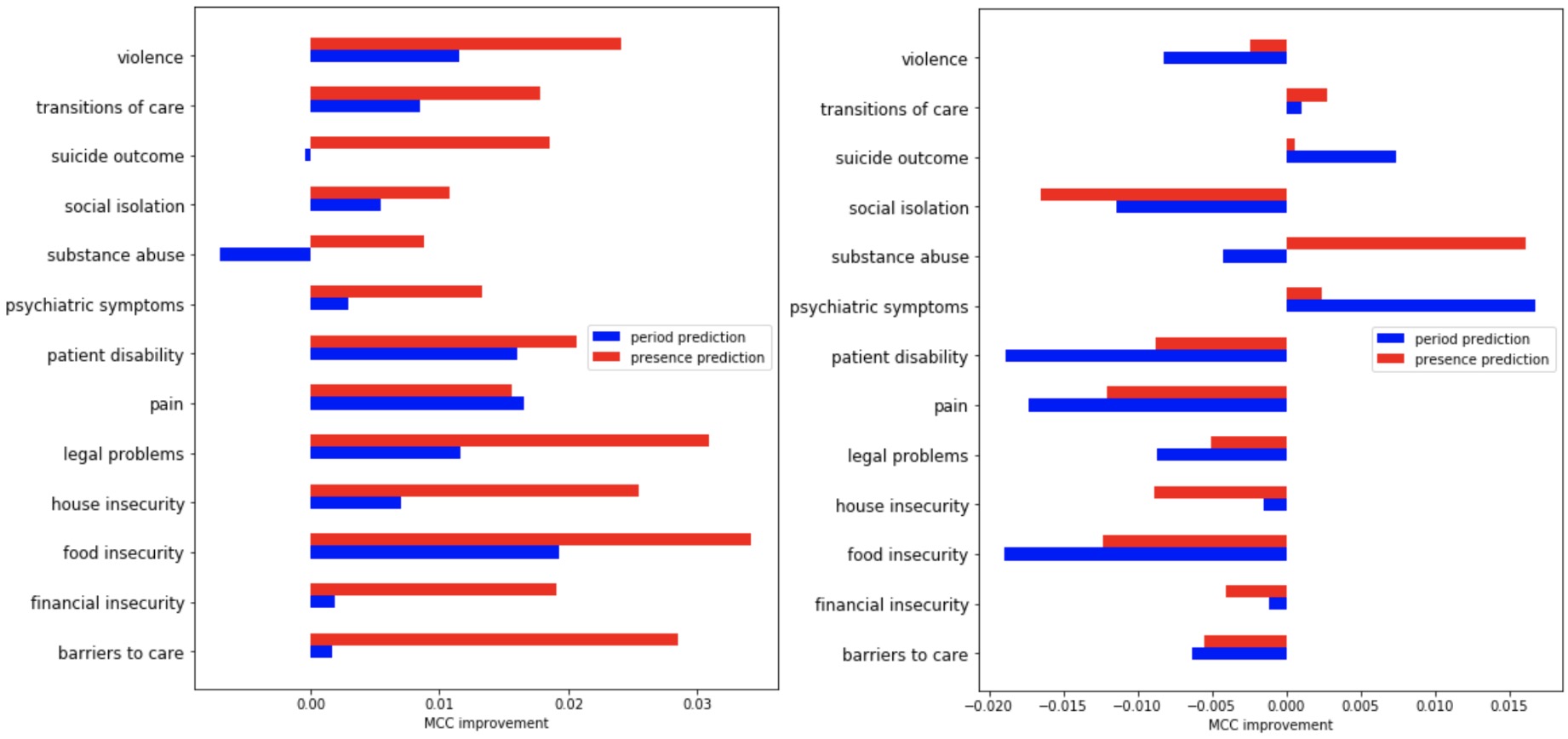}
        \caption{MCC difference before and after adding specific SBDH for Bio\_ClinicalBERT (left), 
        and MCC difference before and after deleting specific SBDH from Bio\_ClinicalBERT-KIRESH (right).
        Here, we try to explore the Ripple Effects of other SBDH for the eviction status classification task.}
        \label{fig:MCC_difference}
    \end{figure*}

According to the Ripple Effects of SBDH \cite{janelle2018social}, no single SBDH is solely responsible for people's situations. Instead, different SBDH form a complex network of causes and effects, and like an actual spider web, pulling on a single point or connection creates ripple effects throughout the web. Our KIRESH model is based on this ripple effect theory, and the results support the effectiveness of the KIRESH model.

As shown in Table \ref{Table:main-results}, KIRESH substantially improved both BioBERT and Bio\_ClinicalBERT.
Note that our current KIRESH model is limited as it depends on an NLP system to extract other SBDH information.  
The NLP system introduces noise and moreover, since it only processes at a sentence level, it may miss other SBDH described elsewhere. 
Despite the aforementioned limitations, the improvement in our implemented KIRESH model is substantial. A future direction to improve KIRESH is to incorporate further other "missing" SBDH information, which may be identified using patient-level structured data of SBDH and NLP extraction of SBDH using patients' longitudinal EHRs.  

We further examined specific SBDH contributing to eviction status classification using a permutation strategy. 
Figure \ref{fig:MCC_difference} show two sets of experimental results.
In the first set of experiments, as shown in Figure \ref{fig:MCC_difference}, we use MCC as an indicator to observe the difference in model performance when we add a certain SBDH as knowledge to Bio\_ClinicalBERT. 
The results show that every social determinant of health factors helps improve the eviction status classification. 
In contrast, behavioral determinants of health (substance abuse and suicidal behaviors) contributed to model performance decline. 
The MCC score change in our permutation experiment also aligns with the aforementioned findings. 
The results, as shown on Figure \ref{fig:MCC_difference}, show that the performance of Bio\_ClinicalBERT-KIRESH  increased after removing psychiatric symptoms, substance abuse, suicide behaviors, and transitions of care from the knowledge. 
This means that these four SBDH actually bring more noise than useful information for the model prediction, or the information network formed by other SBDH is more accurate.
Among the remaining SBDH, food insecurity, pain, patient disability, and social isolation are the SBDH that contribute the most to the Bio\_ClinicalBERT-KIRESH.

One possible interpretation is that while individual social determinants of health (e.g., food insecurity, financial insecurity, and social isolation) and eviction status could be acute events, substance use, psychiatric symptoms, and suicidal behaviors, with the exception of overdose and suicide attempt and death, can be long lasting. In addition, the prevalence of mental health is high among Veterans, which may lead to a weak impact.

Intuitively, the relationship between housing instability (HI) and eviction should be the closest. However, as shown in Figure \ref{fig:MCC_difference}, the impact of HI ranks 4th, which is lower than food insecurity, legal problems, and barriers to care.

\subsubsection*{How does KIRESH-Prompt perform in different data sizes?}
As shown in Table \ref{Table:main-results}, KIRESH-Prompt performed the best when the training size was sufficient. 
Specifically, when the amount of training data is small (10\% and 20\% of the total available training data), our KIRESH-Prompt under-performed the Bio\_ClinicalBERT's fine-tuning model. 
When the training data is 30\% or higher, KIRESH-Prompt shows improvement across most of the evaluation metrics. 
The results are not surprising since other SBDH information also post a data sparsity challenge: since the NLP system outputs 13 SBDH categories, our model KIRESH-Prompt needs sufficient training data to learn the Ripper Effect. 
One way to mitigate this challenge is to incorporate SBDH that could be captured from additional data resources, as we previously mentioned longitudinal and structured data.

\subsection*{Why does our designed prompt work well for Eviction Presence and Period Predictions?}
We hypothesized that the designed prompt could make good use of the intrinsic connection between eviction presence prediction and eviction period prediction.
In the experiment, we find that if the eviction period is already known, the MCC, Macro-F1, and Micro-F1 of the eviction presence prediction will rise from about 0.65, 0.6, and 0.77 to 0.8, 0.7, and 0.8, respectively. If the eviction presence is already known, then the MCC, Macro-F1, and Micro-F1 of the eviction period prediction will rise from about 0.71, 0.69, and 0.82 to 0.92, 0.87, and 0.95, respectively.
Therefore, in the prompt "The eviction presence is [MASK] . The eviction period is [MASK] .", it is much more difficult for the model to predict two [MASK] at the same time than to replace one of the [MASK] with the corresponding label and only predict the other one. When we input our designed prompt into the model, the training becomes more stable because the model is exposed to the same data at varying levels of difficulty.
When we input our designed prompt into the model, the training becomes more stable because the model is exposed to the same data at varying levels of difficulty.
Because when the model makes mistakes without any label in the prompt, it can know more information through the other two simpler data, which can also be regarded as self-reasoning in training.
For example, the model predicts "present" and "current" because they appeared frequently in the previous training, but the correct results were "pending" and "future".
But in fact, if the model already correctly predicts "pending", it can also correctly predict "future", and vice versa. 
So some errors come from the influence of the model on the misprediction of the other task, so the error accumulates in this process.
If we use "The eviction presence is [pending] . The eviction period is [MASK] ." and "The eviction presence is [MASK] . The eviction period is [future] ." to tell the model the logic of the correct prediction, the model can more stably update the weights in the correct direction during training.

\subsection*{Can we trust our models' prediction?}
PLMs always have impressive performance on various tasks, but people often ignore the over-confidence issues that it is likely to appear. In this paper, we want to make our eviction status prediction models as well-calibrated as possible while maintaining high metrics scores. In Table \ref{Table:Calibration}, we use the ECE score to evaluate different methods' calibration. 
Whether a model has a good ECE score but performs poorly on other performance metrics (such as logistic regression), or a model has good accuracy on the dataset, but the ECE score is bad (such as Bio\_ClinicalBERT-KIRESH-Prompt), we will not want to use it in real problems. The former predictions are not convincing, while the latter predictions are likely to be strongly biased. Therefore, the models in Table \ref{Table:main-results} cannot be deployed to real-world problems without calibration.
Temperature scaling is the easiest, fastest, and most straightforward calibration method. And, it's often the most effective and widely used method \cite{guo2017calibration}. In Table \ref{Table:Calibration}, Temperature scaling greatly improves each method's ECE scores without affecting metrics like MCC or F1. This ensures that we can rely on the predictions of the final model after calibration and deploy it to the VHA surveillance system in the future.


\subsection*{Generalizability for the Methods and Analysis in our Research}
The VA population does not represent the general US population. However, many studies and innovations from the VHA have been shown to assist non-VHA facilities in adopting better clinical practices \cite{blosnich2020social, Mitra2022SBDHandSuicide, oliver2008public, hicken2010model, fihn2014insights}. 
In addition, in order to increase the reproduction of our methods, related experiments were conducted and replicated using our applied methods on a publicly available MIMIC3-SBDH dataset. We put the results in Appendix and released the code on GitHub \footnote{https://github.com/seasonyao/KIRESH-Prompt-}

\subsection*{Limitations 
} 
We used a list of keywords (e.g., notice to pay rent, notice to vacate) to select notes for eviction status annotation. This data sampling technique will enrich the dataset for eviction-related notes, which is great for cost-effective annotation. On the other hand, using keyword-based filtering can potentially miss some less-obvious ways of describing evictions. On the other hand, information redundancy, namely that the same eviction status may be stated multiple times throughout an EHR note using varied natural language expressions, our comprehensive list of keywords, which have been carefully selected by domain experts, may mitigate such a bias.


\section*{Conclusion}
In this study, we first created a taxonomy of eviction statuses and then chart-reviewed 5000 EHR notes of the VA health care data to annotate a total of 5,271 instances of eviction status. We then developed an NLP system to automatically detect mentions of eviction status from EHR notes, which combined three simple yet effective methods: KIRESH, Prompt Design, and Temperature Scaling. Our experimental results show our proposed methods substantially improved eviction status classification. In future work, we will evaluate the generalizability of the model framework to other applications.

\section*{Acknowledgments}
We thank our annotators from the Center for Healthcare Organization \& Implementation Research, Veterans Affairs Bedford Healthcare System, Bedford, Massachusetts, for annotating electronic health record notes that were essential for training our natural language processing system.

\section*{FUNDING}
Research reported in this study was supported by the National Institute of Nursing Research and the National Institute of Mental Health of the National Institutes of Health under award numbers 1R01NR020868 and R01MH125027, respectively. This work was supported with funding by the National Center on Homelessness among Veterans, U.S. Department of Veterans Affairs Homeless Programs Office. The content is solely the responsibility of the authors and does not necessarily represent NIMH, NINR, NIH, US Department of Veterans Affairs, or the US government.


\section*{AUTHOR CONTRIBUTIONS}
David A. Levy and Emily Druhl are annotators who created an eviction status taxonomy and annotated 5,271 eviction instances in the Veterans Health Administration. They also participated in paper writing.

Jack Tsai is the domain expert for the two SBDHs of Homelessness and Evictions. He helped write the project proposal with Hong Yu, design the annotation guideline with David A. Levy and Emily Druhl, and also analyzed the final results from his domain knowledge.

Weisong Liu built the annotation tool and was responsible for selecting suitable notes for annotation from a large number of VHA databases. He was also responsible for the subsequent deployment of the model trained in this paper to the VHA as a surveillance system.

Joel I Reisman participated in the formulation and discussion of the project as an experienced healthcare researcher. He provided many valuable views and suggestions from the perspective of healthcare research.

Hong Yu is the planner of the whole project. She was responsible for writing the proposal for the project, planning the direction and progress of the whole project, discussing and suggesting the establishment of the NLP system, and writing the final paper together with Zonghai Yao.

Zonghai Yao is the implementer who built the NLP system. He was responsible for building and analyzing various NLP models and methods proposed in this paper, and together with Hong Yu as the main writer of the final paper.

\section*{CONFLICT OF INTEREST STATEMENT}
No, there are no competing interests

\section*{DATA AVAILABILITY}
The eviction data underlying this article are not publicly available.
The MIMIC-SBDH data underlying this article is a publicly available dataset online \cite{ahsan2021mimic}, which can be found here https://github.com/hibaahsan/MIMIC-SBDH.

\makeatletter
\renewcommand{\@biblabel}[1]{\hfill #1.}
\makeatother

\bibliographystyle{vancouver}
\bibliography{amia}

\begin{thebibliography}{10}

\bibitem{desmond2016evicted}
Desmond M.
\newblock Evicted: Poverty and profit in the American city.
\newblock Crown; 2016.

\bibitem{tsai2019systematic}
Tsai J, Huang M.
\newblock Systematic review of psychosocial factors associated with evictions.
\newblock Health \& Social Care in the Community. 2019;27(3):e1-9.

\bibitem{desmond2015eviction}
Desmond M, Kimbro RT.
\newblock Eviction's fallout: housing, hardship, and health.
\newblock Social forces. 2015;94(1):295-324.

\bibitem{cutts2011us}
Cutts DB, Meyers AF, Black MM, Casey PH, Chilton M, Cook JT, et~al.
\newblock US housing insecurity and the health of very young children.
\newblock American journal of public health. 2011;101(8):1508-14.

\bibitem{hwang2001homelessness}
Hwang SW.
\newblock Homelessness and health.
\newblock Cmaj. 2001;164(2):229-33.

\bibitem{chen2019biosentvec}
Chen Q, Peng Y, Lu Z.
\newblock BioSentVec: creating sentence embeddings for biomedical texts.
\newblock In: 2019 IEEE International Conference on Healthcare Informatics
  (ICHI). IEEE; 2019. p. 1-5.

\bibitem{cox1958regression}
Cox DR.
\newblock The regression analysis of binary sequences.
\newblock Journal of the Royal Statistical Society: Series B (Methodological).
  1958;20(2):215-32.

\bibitem{lee2020biobert}
Lee J, Yoon W, Kim S, Kim D, Kim S, So CH, et~al.
\newblock BioBERT: a pre-trained biomedical language representation model for
  biomedical text mining.
\newblock Bioinformatics. 2020;36(4):1234-40.

\bibitem{alsentzer2019publicly}
Alsentzer E, Murphy JR, Boag W, Weng WH, Jin D, Naumann T, et~al.
\newblock Publicly available clinical BERT embeddings.
\newblock arXiv preprint arXiv:190403323. 2019.

\bibitem{janelle2018social}
Schrag J. Social Determinants of Health: The Ripple Effect; 2018.
\newblock Available from:
  \url{https://essentialhospitals.org/institute/social-determinants-of-health-the-ripple-effect/#:~:text=Each\%20determinant\%20touches\%20health\%20in,being\%20unemployed\%20and\%2For\%20homeless.}

\bibitem{Mitra2022SBDHandSuicide}
Mitra A, Pradhan R, Melamed RD, Chen K, Hoaglin DC, Tucker KL, et~al.
\newblock {Associations Between Natural Language Processing–Enriched Social
  Determinants of Health and Suicide Death Among US Veterans}.
\newblock JAMA Network Open. 2023 03;6(3):e233079-9.
\newblock Available from:
  \url{https://doi.org/10.1001/jamanetworkopen.2023.3079}.

\bibitem{uzuner2008identifying}
Uzuner {\"O}, Goldstein I, Luo Y, Kohane I.
\newblock Identifying patient smoking status from medical discharge records.
\newblock Journal of the American Medical Informatics Association.
  2008;15(1):14-24.

\bibitem{alzoubi2018automated}
Alzoubi H, Ramzan N, Alzubi R, Mesbahi E.
\newblock An automated system for identifying alcohol use status from clinical
  text.
\newblock In: 2018 International Conference on Computing, Electronics \&
  Communications Engineering (iCCECE). IEEE; 2018. p. 41-6.

\bibitem{wicentowski2008using}
Wicentowski R, Sydes MR.
\newblock Using implicit information to identify smoking status in smoke-blind
  medical discharge summaries.
\newblock Journal of the American Medical Informatics Association.
  2008;15(1):29-31.

\bibitem{carrero2006quick}
Carrero F, Hidalgo J, Puertas E, Ma{\~n}a M, Mata J.
\newblock Quick prototyping of high performance text classifiers.
\newblock In: i2b2 Workshop on Challenges in Natural Language Processing for
  Clinical Data; 2006. .

\bibitem{pedersen2006determining}
Pedersen T.
\newblock Determining smoker status using supervised and unsupervised learning
  with lexical features.
\newblock In: i2b2 Workshop on Challenges in Natural Language Processing for
  Clinical Data. Citeseer; 2006. .

\bibitem{jonnagaddala2015preliminary}
Jonnagaddala J, Dai HJ, Ray P, Liaw ST.
\newblock A preliminary study on automatic identification of patient smoking
  status in unstructured electronic health records.
\newblock In: Proceedings of BioNLP 15; 2015. p. 147-51.

\bibitem{topaz2019extracting}
Topaz M, Murga L, Bar-Bachar O, Cato K, Collins S.
\newblock Extracting alcohol and substance abuse status from clinical notes:
  the added value of nursing data.
\newblock In: MEDINFO 2019: Health and Wellbeing e-Networks for All. IOS Press;
  2019. p. 1056-60.

\bibitem{mikolov2013efficient}
Mikolov T, Chen K, Corrado G, Dean J.
\newblock Efficient estimation of word representations in vector space.
\newblock arXiv preprint arXiv:13013781. 2013.

\bibitem{wu2020phrase2vec}
Wu Y, Zhao S, Li W.
\newblock Phrase2Vec: phrase embedding based on parsing.
\newblock Information Sciences. 2020;517:100-27.

\bibitem{gundlapalli2013using}
Gundlapalli AV, Carter ME, Palmer M, Ginter T, Redd A, Pickard S, et~al.
\newblock Using natural language processing on the free text of clinical
  documents to screen for evidence of homelessness among US veterans.
\newblock In: AMIA Annual Symposium Proceedings. vol. 2013. American Medical
  Informatics Association; 2013. p. 537.

\bibitem{ahsan2021mimic}
Ahsan H, Ohnuki E, Mitra A, You H.
\newblock MIMIC-SBDH: A Dataset for Social and Behavioral Determinants of
  Health.
\newblock In: Machine Learning for Healthcare Conference. PMLR; 2021. p.
  391-413.

\bibitem{vasquez2017threat}
V{\'a}squez-Vera H, Pal{\`e}ncia L, Magna I, Mena C, Neira J, Borrell C.
\newblock The threat of home eviction and its effects on health through the
  equity lens: a systematic review.
\newblock Social science \& medicine. 2017;175:199-208.

\bibitem{gundlapalli2014extracting}
Gundlapalli AV, Carter ME, Divita G, Shen S, Palmer M, South B, et~al.
\newblock Extracting concepts related to homelessness from the free text of VA
  electronic medical records.
\newblock In: AMIA Annual Symposium Proceedings. vol. 2014. American Medical
  Informatics Association; 2014. p. 589.

\bibitem{fetzer2019housing}
Fetzer T, Sen S, Souza PC.
\newblock Housing insecurity, homelessness and populism: Evidence from the UK.
\newblock SSRN. 2019.

\bibitem{treglia2023quantifying}
Treglia D, Byrne T, Tamla~Rai V.
\newblock Quantifying the Impact of Evictions and Eviction Filings on
  Homelessness Rates in the United States.
\newblock Housing Policy Debate. 2023:1-12.

\bibitem{kotsiantis2007supervised}
Kotsiantis SB, Zaharakis I, Pintelas P, et~al.
\newblock Supervised machine learning: A review of classification techniques.
\newblock Emerging artificial intelligence applications in computer
  engineering. 2007;160(1):3-24.

\bibitem{radford2018improving}
Radford A, Narasimhan K, Salimans T, Sutskever I, et~al.
\newblock Improving language understanding by generative pre-training.
\newblock arXiv preprint. 2018.

\bibitem{peters-etal-2018-deep}
Peters ME, Neumann M, Iyyer M, Gardner M, Clark C, Lee K, et~al.
\newblock Deep Contextualized Word Representations.
\newblock In: Proceedings of the 2018 Conference of the North {A}merican
  Chapter of the Association for Computational Linguistics: Human Language
  Technologies, Volume 1 (Long Papers). New Orleans, Louisiana: Association for
  Computational Linguistics; 2018. p. 2227-37.
\newblock Available from: \url{https://aclanthology.org/N18-1202}.

\bibitem{devlin2018bert}
Devlin J, Chang MW, Lee K, Toutanova K.
\newblock Bert: Pre-training of deep bidirectional transformers for language
  understanding.
\newblock arXiv preprint arXiv:181004805. 2018.

\bibitem{kwon2022medjex}
Kwon S, Yao Z, Jordan HS, Levy DA, Corner B, Yu H.
\newblock MedJEx: A Medical Jargon Extraction Model with Wiki's Hyperlink Span
  and Contextualized Masked Language Model Score.
\newblock arXiv preprint arXiv:221005875. 2022.

\bibitem{yao2020zero}
Yao Z, Cao L, Pan H.
\newblock Zero-shot entity linking with efficient long range sequence modeling.
\newblock arXiv preprint arXiv:201006065. 2020.

\bibitem{guo2017calibration}
Guo C, Pleiss G, Sun Y, Weinberger KQ.
\newblock On calibration of modern neural networks.
\newblock In: International conference on machine learning. PMLR; 2017. p.
  1321-30.

\bibitem{jiang2021can}
Jiang Z, Araki J, Ding H, Neubig G.
\newblock How can we know when language models know? on the calibration of
  language models for question answering.
\newblock Transactions of the Association for Computational Linguistics.
  2021;9:962-77.

\bibitem{frenkel2021network}
Frenkel L, Goldberger J.
\newblock Network Calibration by Class-based Temperature Scaling.
\newblock In: 2021 29th European Signal Processing Conference (EUSIPCO). IEEE;
  2021. p. 1486-90.

\bibitem{gorodkin2004comparing}
Gorodkin J.
\newblock Comparing two K-category assignments by a K-category correlation
  coefficient.
\newblock Computational biology and chemistry. 2004;28(5-6):367-74.

\bibitem{naeini2015obtaining}
Naeini MP, Cooper G, Hauskrecht M.
\newblock Obtaining well calibrated probabilities using bayesian binning.
\newblock In: Twenty-Ninth AAAI Conference on Artificial Intelligence; 2015. .

\bibitem{blosnich2020social}
Blosnich JR, Montgomery AE, Dichter ME, Gordon AJ, Kavalieratos D, Taylor L,
  et~al.
\newblock Social determinants and military veterans’ suicide ideation and
  attempt: a cross-sectional analysis of electronic health record data.
\newblock Journal of general internal medicine. 2020;35:1759-67.

\bibitem{oliver2008public}
Oliver A.
\newblock Public-sector health-care reforms that work? A case study of the US
  Veterans Health Administration.
\newblock The lancet. 2008;371(9619):1211-3.

\bibitem{hicken2010model}
Hicken BL, Plowhead A.
\newblock A model for home-based psychology from the veterans health
  administration.
\newblock Professional Psychology: Research and Practice. 2010;41(4):340.

\bibitem{fihn2014insights}
Fihn SD, Francis J, Clancy C, Nielson C, Nelson K, Rumsfeld J, et~al.
\newblock Insights from advanced analytics at the Veterans Health
  Administration.
\newblock Health affairs. 2014;33(7):1203-11.

\bibitem{department2012national}
Department~of Veterans~Affairs U, et~al.. National center for Veteran’s
  analysis and statistics. Washington, DC: Department of Veterans Affairs.
  Retrieved from http://www~…; 2012.

\bibitem{grandini2020metrics}
Grandini M, Bagli E, Visani G.
\newblock Metrics for multi-class classification: an overview.
\newblock arXiv preprint arXiv:200805756. 2020.

\bibitem{hill2021social}
Hill-Briggs F, Adler NE, Berkowitz SA, Chin MH, Gary-Webb TL, Navas-Acien A,
  et~al.
\newblock Social determinants of health and diabetes: a scientific review.
\newblock Diabetes care. 2021;44(1):258-79.

\bibitem{nijhawan2019clinical}
Nijhawan AE, Metsch LR, Zhang S, Feaster DJ, Gooden L, Jain MK, et~al.
\newblock Clinical and sociobehavioral prediction model of 30-day hospital
  readmissions among people with HIV and substance use disorder: beyond
  electronic health record data.
\newblock Journal of acquired immune deficiency syndromes (1999).
  2019;80(3):330.

\bibitem{takahashi2015health}
Takahashi PY, Ryu E, Olson JE, Winkler EM, Hathcock MA, Gupta R, et~al.
\newblock Health behaviors and quality of life predictors for risk of
  hospitalization in an electronic health record-linked biobank.
\newblock International journal of general medicine. 2015:247-54.

\bibitem{zheng2020development}
Zheng L, Wang O, Hao S, Ye C, Liu M, Xia M, et~al.
\newblock Development of an early-warning system for high-risk patients for
  suicide attempt using deep learning and electronic health records.
\newblock Translational psychiatry. 2020;10(1):72.

\bibitem{haas2015proactive}
Haas JS, Linder JA, Park ER, Gonzalez I, Rigotti NA, Klinger EV, et~al.
\newblock Proactive tobacco cessation outreach to smokers of low socioeconomic
  status: a randomized clinical trial.
\newblock JAMA internal medicine. 2015;175(2):218-26.

\bibitem{hamilton2012barriers}
Hamilton AB, Poza I, Hines V, Washington DL.
\newblock Barriers to psychosocial services among homeless women veterans.
\newblock Journal of Social Work Practice in the Addictions. 2012;12(1):52-68.

\bibitem{dorr2019identifying}
Dorr D, Bejan CA, Pizzimenti C, Singh S, Storer M, Quinones A.
\newblock Identifying patients with significant problems related to social
  determinants of health with natural language processing.
\newblock In: MEDINFO 2019: Health and Wellbeing e-Networks for All. IOS Press;
  2019. p. 1456-7.

\bibitem{feller2020detecting}
Feller DJ, Don't~Walk OJB, Zucker J, Yin MT, Gordon P, Elhadad N, et~al.
\newblock Detecting social and behavioral determinants of health with
  structured and free-text clinical data.
\newblock Applied clinical informatics. 2020;11(01):172-81.

\bibitem{wang-etal-2019-harnessing}
Wang Y, Wu Y, Mou L, Li Z, Chao W.
\newblock Harnessing Pre-Trained Neural Networks with Rules for Formality Style
  Transfer.
\newblock In: Proceedings of the 2019 Conference on Empirical Methods in
  Natural Language Processing and the 9th International Joint Conference on
  Natural Language Processing (EMNLP-IJCNLP). Hong Kong, China: Association for
  Computational Linguistics; 2019. p. 3573-8.
\newblock Available from: \url{https://aclanthology.org/D19-1365}.

\bibitem{yao2021improving}
Yao Z, Yu H.
\newblock Improving formality style transfer with context-aware rule injection.
\newblock arXiv preprint arXiv:210600210. 2021.

\bibitem{yin-etal-2019-benchmarking}
Yin W, Hay J, Roth D.
\newblock Benchmarking Zero-shot Text Classification: Datasets, Evaluation and
  Entailment Approach.
\newblock In: Proceedings of the 2019 Conference on Empirical Methods in
  Natural Language Processing and the 9th International Joint Conference on
  Natural Language Processing (EMNLP-IJCNLP). Hong Kong, China: Association for
  Computational Linguistics; 2019. p. 3914-23.
\newblock Available from: \url{https://aclanthology.org/D19-1404}.

\bibitem{yang2022knowledge}
Yang Z, Wang S, Rawat BPS, Mitra A, Yu H.
\newblock Knowledge Injected Prompt Based Fine-tuning for Multi-label Few-shot
  {ICD} Coding.
\newblock In: Findings of the Association for Computational Linguistics: EMNLP
  2022. Abu Dhabi, United Arab Emirates: Association for Computational
  Linguistics; 2022. p. 1767-81.
\newblock Available from:
  \url{https://aclanthology.org/2022.findings-emnlp.127}.

\bibitem{liu2021pre}
Liu P, Yuan W, Fu J, Jiang Z, Hayashi H, Neubig G.
\newblock Pre-train, prompt, and predict: A systematic survey of prompting
  methods in natural language processing.
\newblock arXiv preprint arXiv:210713586. 2021.

\bibitem{gao-etal-2021-making}
Gao T, Fisch A, Chen D.
\newblock Making Pre-trained Language Models Better Few-shot Learners.
\newblock In: Proceedings of the 59th Annual Meeting of the Association for
  Computational Linguistics and the 11th International Joint Conference on
  Natural Language Processing (Volume 1: Long Papers). Online: Association for
  Computational Linguistics; 2021. p. 3816-30.
\newblock Available from: \url{https://aclanthology.org/2021.acl-long.295}.

\bibitem{yao2022extracting}
Yao Z, Cao Y, Yang Z, Deshpande V, Yu H.
\newblock Extracting Biomedical Factual Knowledge Using Pretrained Language
  Model and Electronic Health Record Context.
\newblock arXiv preprint arXiv:220907859. 2022.

\bibitem{yao2022context}
Yao Z, Cao Y, Yang Z, Yu H.
\newblock Context Variance Evaluation of Pretrained Language Models for
  Prompt-based Biomedical Knowledge Probing.
\newblock arXiv preprint arXiv:221110265. 2022.

\bibitem{johnson2016mimic}
Johnson AE, Pollard TJ, Shen L, Lehman LwH, Feng M, Ghassemi M, et~al.
\newblock MIMIC-III, a freely accessible critical care database.
\newblock Scientific data. 2016;3(1):1-9.

\end{thebibliography}

\newpage
\appendix
\section*{Appendix}
\label{appendix:appendix}

\subsection*{Appendix-A Annotation Guideline}
\label{appendix:anno-example}

\paragraph{Objectives of Study} 
Our project aims to develop NLP models that can automatically detect incidence of eviction in Veterans’ EHR notes

\paragraph{Data Source}
This study used the EHR database from the VHA Corporate Data Warehouse (CDW). With a primary obligation to provide medical services to all eligible US Veterans, the VHA is the largest integrated healthcare network in the country; its EHR system spans more than a thousand medical centers and clinics \cite{department2012national}. The VHA database includes patient demographic information, medication, diagnoses, procedures, clinical notes, and billing. Our study protocol was approved by the institutional review board of US Veterans Affairs (VA) Bedford Health Care, and we obtained a waiver of informed consent. 

\paragraph{Annotation Schema}
We identified \emph{eviction presence} (Is eviction/quit indicated in the EHR?) and \emph{eviction period} (Is the eviction/quit current?). Presence includes Present, Absent, Pending, Uncertain, Hypothetical, Quit/mutual termination [[technically a separate class from eviction]], Irrelevant (context not related to Veteran); if used, period is automatically irrelevant also. Period includes Current (present tense in EHR note), Future, Uncertain (cannot be determined by context). Along with eviction, we also captured instances of mutual recission between Veterans and landlords, which we labeled as ``quit’' to differentiate from formal eviction. When eviction context does not directly involve the Veteran’s own housing status, we labeled as ``irrelevant’'.

\subsection*{Rules}

1. When ``eviction’' is used with no other context, annotate as current/pending.

2. Use tense as indicator of period:
``being evicted’' = current/present
``was evicted’' = history/present

3. If a date is given (ex. he must pay rent by 11/3 or be evicted on 11/9), annotate as current/present. 

4. Receipt of any type of letter – intent to evict, 3 or 5 or 10 day notices, letter of eviction, etc. constitutes eviction process has at least begun; annotate current/present.

\subsection*{Examples}

\begin{enumerate}
    \item History/absent
        \begin{enumerate}
            \item "background check shows Veteran was never evicted"
        \end{enumerate}
    \item History/present
        \begin{enumerate}
            \item "has been evicted"
            \item "was evicted from previous residence"
        \end{enumerate}
    \item History/uncertain
        \begin{enumerate}
            \item "History/uncertain"
        \end{enumerate}
    \item Current/absent
        \begin{enumerate}
            \item "landlord denies they are being evicted"
            \item "landlord explained she didn’t intend to evict him"
            \item "agreed not to evict him"
            \item "eviction has been rescinded" 
        \end{enumerate}
        Current/absent with language suggesting eviction is only presently avoided turns into future/hypothetical: "landlord says he will not evict at this time"
     \item Current/present
        \begin{enumerate}
            \item "is getting/being evicted"
            \item "got eviction notice today"
            \item "received an eviction notice"/"notice of eviction"
            \item "received eviction notice and must be out by 5th February"
            \item "sent a letter of intent to evict" (any step in the process from when the first letter is sent)
            \item "landlord is pursuing eviction"
        \end{enumerate}
        A future/hypothetical with a set timeline: "he may get evicted in two days"
    \item Current/pending (steps have already been taken/decision has been made to evict but written notice has not been given to patient)
        \begin{enumerate}
            \item "planning to evict"
            \item "trying to evict him" 
            \item "intend to evict"
            \item "landlord has an eviction notice to serve her"
            \item "I am about to get evicted"
            \item "facing eviction"
            \item "threatening to evict"
            \item "expected to be evicted"
            \item "likely to be evicted soon"
            \item "in imminent danger of being evicted"
            \item "He is one month behind on his rent and was given a verbal notice to evict"
        \end{enumerate}
    \item Current/uncertain (current but not enough context to determine whether present/pending/absent)
        \begin{enumerate}
            \item "This SW phoned patient re: eviction"
            \item "if patient had been issued an eviction notice"
            \item "pt did not show up to eviction hearing" – eviction hearings can go either way
            \item "f/u regarding possible eviction"
            \item "they can evict me, I don’t care"
            \item "no forward motion on eviction yet"
        \end{enumerate}
    \item Future/hypothetical (any context that suggests the decision to evict has not yet been made  (if/then statements) or the patient voices concerns regarding a potential eviction)
        \begin{enumerate}
            \item "possibly facing eviction"
            \item "may be evicted"
            \item "in danger of being evicted"
            \item "requested paperwork for eviction prevention support"
            \item "afraid of an eviction"
            \item "can’t evict her during the pandemic"
            \item "indicated that he won’t evict at this time"
            \item "so he doesn’t get evicted"
            \item "if landlord moves forward with eviction"
            \item "LL willing to work with them to avoid eviction"
        \end{enumerate}
    \item Irrelevant
        \begin{enumerate}
            \item "patient has to evict his son"
            \item "Veteran’s girlfriend is getting evicted"
        \end{enumerate}
    \item MR /present
        \begin{enumerate}
            \item If notice sent/decision made
            \item "Landlord and Veteran agreed on mutual rescission at end of month"
        \end{enumerate}
    \item MR /pending
        \begin{enumerate}
            \item Notice not yet sent
            \item "Landlord states will talk with Veteran about mutual rescission instead"
        \end{enumerate}

\end{enumerate}

    \begin{table*}
        \centering
        \begin{tabularx}{0.3\textwidth}{l|c}
        \hline
        \footnotesize{Categoty} & \footnotesize{Fleiss’ kappa}
        \\
        
        \hline
        \footnotesize{History/absent} & \footnotesize{1}
        \\
        \footnotesize{History/present} & \footnotesize{0.9989}
        \\
        \footnotesize{History/uncertain} & \footnotesize{1} 
        \\
        \footnotesize{History/quit} & \footnotesize{1}
        \\
        \footnotesize{Current/absent} & \footnotesize{0.9735}
        \\
        \footnotesize{Current/present} & \footnotesize{0.9947}
        \\
        \footnotesize{Current/pending} & \footnotesize{0.9884}
        \\
        \footnotesize{Current/uncertain} & \footnotesize{0.9759}
        \\
        \hline
        \footnotesize{Future/hypothetical} & \footnotesize{0.9975}
        \\
        \footnotesize{Future/quit} & \footnotesize{1}
        \\
        \hline
        \footnotesize{Irrelevant} & \footnotesize{1}
        \\
        \hline
        \footnotesize{Quit/current} & \footnotesize{0}
        \\
        \footnotesize{Quit/present} & \footnotesize{0.9844}
        \\
        \footnotesize{Quit/pending} & \footnotesize{0.9695}
        \\
        \footnotesize{Uncertain/present} & \footnotesize{0.9865}
        \\
        \footnotesize{Uncertain/uncertain} & \footnotesize{0.9227}
        \\
        \footnotesize{Avg} & \footnotesize{0.9289}
        \\

        \hline
        \end{tabularx}
        \caption{The results showed that there was good agreement among annotators (Fleiss’ kappa).
        } 
        \label{data:Fleiss-kappa}
    \end{table*}

\subsection*{Appendix-B Experimental Setting}
\label{appendix:setting}

For logistic regression model training, we use default setting for scikit-learn \footnote{https://scikit-learn.org/stable/modules/generated/sklearn.linear\_model.LogisticRegression.html}. 
For fine-tuning, we set hyper-parameters as follows.  Batch size, learning rate, and maximum epoch were set to 8, 1e-5 and 20. There are two new initialized linear layers on the last hidden states of [CLS] token. The linear layer size is 768, and the dropout is 0.3.
During training, cross-entropy loss is utilized.
All experiments were conducted in Ubuntu 20.04 environment using one Tesla P100 GPU, Intel Xeon Gold 6226R CPU.
We set training epoch number to 30, and all experiments can been done within 2 hours.
In addition, we used five different random seeds to sample training data for all our experiments, and the scores reported are the average of these random seeds.

To test our methods' practicality in future real usage when there are billion of notes and sentences, we deploy this model into our Solr service. Here is the performance information: With an NVIDIA A6000 GPU, our pipeline can process about 32 notes per second, including the time to retrieve the notes from our Solr service, process the notes with the model, and save the result to Solr.

\subsection*{Appendix-C Metrics}
\label{appendix:metrics}
Our task is a multi-task learning, where eviction presence and period tasks are both multi-class learning. For model prediction and label pairs, we can calculate the confusion matrix C for both tasks, which are True Positives (TP), True Negatives (TN), False Positives (FP), False Negatives (FN) for each class.
Both tasks are very imbalanced, so we follow the recent work about metrics for multi-class classification \cite{grandini2020metrics} for better evaluation.
First, We use both Macro and Micro methods to average F1 score.
The metrics formulas for eviction presence prediction or period are the same. Use the prediction presence prediction task as an example. For precision the formula is:

\begin{equation}
\begin{aligned}
Macro P = 1/ N_c \sum_{i=1}^{N_c} \frac{TP_i}{TP_i+FP_i} \quad \quad \quad  
Micro P = \frac{\sum_{i=1}^{N_c} TP_i*n_i}{\sum_{i=1}^{N_c} (TP_i+FP_i)*n_i} \\
\end{aligned}
\label{eq:macrop}
\end{equation}

where $N_c$ is the number of classes (6 for eviction presence and 5 for eviction period), and $n_i$ is the number of data for the certain class. For recall the formula is:

\begin{equation}
\begin{aligned}
Macro R = 1/ N_c \sum_{i=1}^{N_c} \frac{TP_i}{TP_i+FN_i} \quad \quad \quad  
Micro R = \frac{\sum_{i=1}^{N_c} TP_i*n_i}{\sum_{i=1}^{N_c} (TP_i+FN_i)*n_i} \\
\end{aligned}
\label{eq:macror}
\end{equation}

So the F1-score according to Equation \ref{eq:macrop} and \ref{eq:macror} will be:

\begin{equation}
\begin{aligned}
Macro-F1 = 2 * \frac{Macro P * Macro R}{Macro P + Macro R} \quad \quad \quad
Micro-F1 = 2 * \frac{Micro P * Micro R}{Micro P + Micro R} \\
\end{aligned}
\label{eq:f1}
\end{equation}

In addition, we also report the Matthews Correlation Coefficient (MCC) score, which uses all the information in the confusion matrix (TP, TN, FP, and FN). The MCC can be defined in terms of a confusion matrix C for K classes \footnote{https://scikit-learn.org/stable/modules/model\_evaluation.html\#matthews-corrcoef}:

\begin{equation}
\begin{aligned}
MCC = \frac{c*s-\sum_{k}^{K}(p_k*t_k)}{\sqrt{(s^2-\sum_{k}^{K}p_k^2)(s^2-\sum_{k}^{K}t_k^2)}}\\
\end{aligned}
\label{eq:MCC}
\end{equation}

To simplify the definition, here we consider the following intermediate variables \cite{gorodkin2004comparing}:

\begin{enumerate}
    \item $c=\sum_{k}^{K}(C_{kk})$ the total number of elements correctly predicted
    \item $s=\sum_{i}^{K}\sum_{j}^{K}(C_{ij})$ the total number of samples.
    \item $t_k=\sum_{i}^{K}(C_{ik})$ the number of times class $k$  truly occurred,
    \item $p_k=\sum_{i}^{K}(C_{ki})$ the number of times class $k$ was predicted,
\end{enumerate}

Finally, we also need a scalar summary statistic of calibration. Expected Calibration Error \cite{guo2017calibration,naeini2015obtaining} – or ECE – partitions predictions into M equally-spaced bins and takes a weighted average of the bins' accuracy/confidence difference. More precisely,

$$ECE = \sum_{m=1}^M \frac{|B_m|}{n}|acc(B_m)-conf(B_m)| $$

where n is the number of samples. The difference between accuracy and confidence for a given bin represents the calibration gap. We use ECE as the primary empirical metric to measure calibration.

\paragraph{Metrics selection} Basically, it's hard to say macro is better or micro is better. And we selected macro because it's friendly to our surveillance system since we care more about the "eviction present" and "current" (majority class) performance. Note that we did not choose the metric by the results, we chose the metric by our final goal, and we reported all the metrics to ensure that different people can have their own evaluation criteria based on different metrics.

\subsection*{Appendix-D More Related Work}
\label{appendix:related-work}

\subsubsection*{Social and Behavioral Determinants of Health}
Social determinants of health (SDOH) are ``the conditions in which people are born, grow, live, work and age’', which are ``shaped by the distribution of money, power and resources’' \cite{hill2021social}. They include factors such as socioeconomic status, education, neighborhood and natural environment, employment, social support networks and access to healthcare. Behavioral determinants of health include smoking, drug use, alcohol use, physical activity and diet. In conclusion, social and behavioral determinants of health (SBDH) are environmental and behavioral factors that have a significant impact on health. Several studies have shown the impact of SBDHs on health outcomes. \cite{nijhawan2019clinical} showed that variables from the SBDH, such as substance use and food access, contribute to a 30-day readmission prediction task. \cite{takahashi2015health} concluded that education level, unemployment status, and alcohol consumption significantly affect hospitalization risk. \cite{zheng2020development} showed that substance abuse, low income, and unemployment can be used to identify high-risk patients for suicide attempts. Because of its importance, physicians record information about a patient's SBDH in an electronic health record (EHR). Information about SBDH is useful to both researchers and clinicians. Researchers study how SBDH affects health outcomes, and clinicians can identify patients at social and behavioral risk to provide tailored care, such as counseling or therapy \cite{haas2015proactive} and social services \cite{hamilton2012barriers}. The challenge for clinicians and researchers is that SBDH is mostly documented in unstructured EHR notes. One study showed that EHR notes contained 90 times more SBDH information than structured EHR data \cite{dorr2019identifying}. However, there is no globally accepted format to record SBDH in EHR notes consistently, and manually viewing notes for SBDH is time-consuming. Therefore, there is a need for automatic identification of patients' SBDH status in EHR notes. Although natural language processing (NLP) methods have been developed to identify the status of several SBDHs in EHR automatically notes \cite{gundlapalli2013using, alzoubi2018automated, feller2020detecting}, the notes and associated annotations. The information used to develop the methods was not made public.

\subsubsection*{Text Classification using NLP and BioNLP}
In NLP and BioNLP domain, the standard of classification tasks have recently shifted to the pre-train and fine-tune paradigm \cite{kotsiantis2007supervised, radford2018improving, peters-etal-2018-deep}.
In this paradigm, language models can be trained on large datasets, in the process learning robust language-related features, and will be then adapted to different downstream tasks by fine-tuning them using task-specific objective functions \cite{devlin2018bert, kwon2022medjex, yao2020zero}.
Many approaches that integrate external knowledge have been developed to overcome the low-resource setting in many downstream tasks \cite{wang-etal-2019-harnessing, yao2021improving}. 
More recently, prompt-based learning has been widely explored in classification-based tasks where prompt templates can be constructed relatively easily \cite{yin-etal-2019-benchmarking, yang2022knowledge}. The key to prompting for classification-based tasks is reformulating it as an appropriate prompt. For example, \cite{yin-etal-2019-benchmarking} uses a prompt such as "the topic of this document is [Z].", which is then fed into language models for slot filling. Prompt-based learning is developed mainly in few-shot or zero-shot settings \cite{liu2021pre, gao-etal-2021-making, yao2022extracting, yao2022context}.
In this paper, our tasks are not the few-shot setting so we followed the pre-train and fine-tune paradigm to train our models. We compared it with the method of pre-trained textual representations. Moreover, our task is done in a low-resource setting. Therefore, we use ripple effects \cite{janelle2018social} of the other 13 SBDH to design the special knowledge for our model, and design a prompt based on the characteristics of eviction presence and period prediction. 

In addition, it is well known that modern neural networks including language models are poorly calibrated, especially in the extremely imbalanced setting \cite{guo2017calibration, jiang2021can, frenkel2021network}. They tend to overestimate or underestimate probabilities when compared to the expected accuracy. This results in misleading reliability and corrupting the decision policy. Temperature scaling is a post-processing method that fixes this \cite{guo2017calibration}. This simple yet effective method does not affect the model’s accuracy. In this paper, we also used temperature scaling for our models in order to avoid over-confidence issues arising from the imbalanced dataset.

\subsection*{Appendix-E Error Analysis in our VA-Eviction dataset}
\subparagraph{SBDH Knowledge Injection}
Noisy and incomplete SBDH knowledge coming from AI will cause some errors. Comparing the results of Bio\_ClinicalBERT and Bio\_ClinicalBERT-KIRESH.
When we injected SBDH knowledge into Bio\_ClinicalBERT, 20.777\% and 16.698\% of the evaluation presence and evaluation period changed from wrong to correct, but at the same time, 19.829\% and 16.034\% of the evaluation presence and evaluation period also changed from correct to wrong. Our analyses also show that training with the SBDH knowledge injection is not stable, further confirming our data sparsity challenge.

\subparagraph{Errors from Imbalanced Data Distribution} 
Data imbalance also contributed to errors.  
If we consider the performance of Bio\_ClinicalBERT-KIRESH-Prompt in different classes. The F1 scores for ``current’', ``future’', ``history’', ``irrelevant’', and ``uncertain’' are 0.86, 0.71, 0.88, 0.82, and 0.11, separately.
And the F1 scores for ``present’', ``absent’', ``pending’', ``Mutual Rescission’', ``uncertain’', and ``no’' are 0.85, 0.62, 0.56, 0.53, 0.29, and 0.78, separately.
Even though we used Temperature Scaling-based for Calibration, it only helps confidence but not real prediction (ECC or F1), so imbalance still influences the prediction. 
So models perform badly in those minority classes like ``uncertain’'.
We will continue to improve this in the future.

\subsection*{Appendix-F More Discussion}
\label{appendix:discussion}

\subsection*{Why is the Pre-training and Fine-tuning Paradigm better?}
Both baseline methods have obtained a lot of domain knowledge through pre-training on PubMed and MIMIC-III, but BioSentV only has relevant knowledge in its embeddings, while PLMs have knowledge in both embeddings and parameters. In addition, fine-tuning the entire PLMs is more effective than just training a logistic regression classifier. Although PLMs use more memory to get a better performance, but Table \ref{Table:main-results} also shows that using only 10\% of the training data (about 300 data), Bio\_ClinicalBERT has greatly surpassed the results of BioSentV and logistic regression, and it is enough for Bio\_ClinicalBERT to use only 20\% of the training data to get a relatively reliable classification performance. Therefore, this strategy of exchanging more memory for less training data is worthwhile for the resource-lean biomedical domain.

\subsubsection*{Why choose the text classification rather than NER?}

This is actually because we are preparing for many subsequent SBDH tasks without sufficient resources. As we all know, the cycle of annotating such a dataset for each SBDH is very long. One of the topics we are studying is how to use continual learning and active learning to help shorten the annotation cycle. However, for CL and AL, the related research on the classification task is significantly more mature than that related to NER. Based on this purpose, we chose the text classification rather than NER for eviction and many other subsequent SBDH tasks. For those spans that are highlighted in the annotation, we will use them as evidence in the classification task in the future.

\subsubsection*{Generalizability of proposed methods}

We conducted our proposed methods on a publicly available MIMIC-SBDH dataset \cite{ahsan2021mimic}. 
We chose MIMIC-SBDH because it is the first publicly available EHR annotation dataset annotating patients' SBDH status. To this end, they annotated 7,025 discharge summaries randomly selected from the MIMIC III \cite{johnson2016mimic} dataset for the following SBDHs: community, economic, educational, environmental, alcohol use, tobacco use, and drug use. In addition, they marked SBDH-related keywords to understand better the language used to discuss them.

However, among all categories, only alcohol use, tobacco use, and drug use have label (Present, Past, Never, Unsure, None) that can be converted into presence and period, and we convert them into two sets of labels, presence the period. If the label is None, then the new label of presence is 1 (None), and the new label of period is 1 (None); if the label is present, the new label of presence is 2 (yes), and the new label of period is 2 (current ); if the label is past, then the new label of presence is 1 (yes), the new label of period is 2 (past); if the label is Never, the new label of presence is 2 (no), and the new label of period is 3 (no); if the label is Unsure, then the new label of presence is 3 (unsure), and the new label of period is 4 (unsure).

We put the results in Table \ref{Table:generalizability} and released the code on GitHub \footnote{https://github.com/seasonyao/KIRESH-Prompt-}.

    \begin{table*}[!htpt]
        \centering
        \begin{tabular}{c|ccc|ccc}
        \hline
        
        & \multicolumn{3}{c|}{eviction period} & \multicolumn{3}{c}{eviction presence} \\
        
        & MCC & Macro-F1 & Micro-F1 & MCC & Macro-F1 & Micro-F1
        \\
        \hline
        
        & \multicolumn{6}{c}{evaluation dataset} \\
        \hline
        
        \scriptsize{BioBERT} & \scriptsize{71.11±0.81} & \scriptsize{68.76±1.12} & \scriptsize{81.32±0.65} & \scriptsize{62.86±0.77} & \scriptsize{56.95±0.86} & \scriptsize{75.85±0.52}
        \\
        
        \scriptsize{BioBERT-KIRESH} & \scriptsize{71.88±1.27} & \scriptsize{68.95±1.63} & \scriptsize{81.75±0.82} & \scriptsize{63.56±1.48} & \scriptsize{57.61±1.61} & \scriptsize{76.22±0.93}
        \\
        
        \scriptsize{BioBERT-Prompt} & \scriptsize{71.68±1.08} & \scriptsize{70.93±1.52} & \scriptsize{81.59±0.61} & \scriptsize{62.89±0.97} & \scriptsize{57.37±1.23} & \scriptsize{75.90±0.82}
        \\
        
        \scriptsize{BioBERT-KIRESH-Prompt} & \scriptsize{73.15±0.74} & \scriptsize{68.93±1.03} & \scriptsize{82.52±0.49} & \scriptsize{63.96±0.79} & \scriptsize{58.38±0.98} & \scriptsize{76.59±0.47}
        \\
        
        \hline
        
        \scriptsize{Bio\_ClinicalBERT} & \scriptsize{72.16±0.75} & \scriptsize{70.98±0.94} & \scriptsize{81.85±0.81} & \scriptsize{64.50±0.64} & \scriptsize{58.45±0.87} & \scriptsize{76.82±0.66}
        
        \\
        
        \scriptsize{Bio\_ClinicalBERT-KIRESH} & \scriptsize{73.16±1.04} & \scriptsize{70.32±1.32} & \scriptsize{82.57±0.88} & \scriptsize{65.88±1.13} & \scriptsize{60.81±1.43} & \scriptsize{77.61±0.91}
        \\
        
        \scriptsize{Bio\_ClinicalBERT-Prompt} & \scriptsize{73.49±1.11} & \scriptsize{69.30±1.17} & \scriptsize{82.86±1.07} & \scriptsize{66.01±0.99} & \scriptsize{61.20±1.03} & \scriptsize{77.73±0.74}
        \\
        
        \scriptsize{Bio\_ClinicalBERT-KIRESH-Prompt} & \scriptsize{74.67±0.83} & \scriptsize{71.15±0.78} & \scriptsize{83.39±0.54} & \scriptsize{66.82±0.63} & \scriptsize{62.73±0.80} & \scriptsize{78.63±0.44}
        \\
        
        \hline
        
        \end{tabular}
        \caption{95\% CI scores for Bio\_ClinicalBERT experimental results.}
        \label{Table:95ci-score-for-main-results}
    \end{table*}

    \begin{table*}[!htpt]
        \centering
        \begin{tabular}{c|ccc|ccc}
        \hline
        
        & \multicolumn{3}{c|}{period} & \multicolumn{3}{c}{presence} \\
        
        & MCC & Macro-F1 & Micro-F1 & MCC & Macro-F1 & Micro-F1
        \\
        \hline
        
        & \multicolumn{6}{c}{Alcohol} \\
        \hline

        \scriptsize{Bio\_ClinicalBERT}  & \scriptsize{84.49±0.36} & \scriptsize{81.56±0.85} & \scriptsize{88.48±0.47} & \scriptsize{86.63±0.37} & \scriptsize{86.66±0.96} & \scriptsize{90.72±0.26}
        
        \\
        
        \scriptsize{Bio\_ClinicalBERT-KIRESH}  & \scriptsize{86.14±1.08} & \scriptsize{82.63±1.82} & \scriptsize{89.63±0.67} & \scriptsize{87.98±0.94} & \scriptsize{87.36±2.20} & \scriptsize{91.54±0.74}
        \\
        
        \scriptsize{Bio\_ClinicalBERT-Prompt}  & \scriptsize{85.13±0.68} & \scriptsize{82.84±1.81} & \scriptsize{89.35±0.62} & \scriptsize{86.43±0.85} & \scriptsize{86.27±1.90} & \scriptsize{90.85±0.59}
        \\
        
        \scriptsize{Bio\_ClinicalBERT-KIRESH-Prompt} & \scriptsize{86.17±0.52} & \scriptsize{83.68±0.82} & \scriptsize{89.65±0.48} & \scriptsize{88.27±0.48} & \scriptsize{87.62±1.51} & \scriptsize{91.85±0.44}
        \\
        
        \hline

        & \multicolumn{6}{c}{Tobacco} \\
        \hline

        \scriptsize{Bio\_ClinicalBERT}  & \scriptsize{85.15±0.55} & \scriptsize{82.22±1.03} & \scriptsize{88.81±0.44} & \scriptsize{87.47±0.53} & \scriptsize{85.44±1.04} & \scriptsize{91.62±0.37}
        
        \\
        
        \scriptsize{Bio\_ClinicalBERT-KIRESH} & \scriptsize{87.04±1.63} & \scriptsize{84.22±1.66} & \scriptsize{90.17±1.41} & \scriptsize{88.52±0.78} & \scriptsize{88.12±1.27} & \scriptsize{92.65±0.72} 
        \\
        
        \scriptsize{Bio\_ClinicalBERT-Prompt} & \scriptsize{86.36±0.75} & \scriptsize{83.70±0.24} & \scriptsize{89.77±0.65} & \scriptsize{87.59±0.45} & \scriptsize{86.58±0.77} & \scriptsize{91.86±0.44} 
        \\
        
        \scriptsize{Bio\_ClinicalBERT-KIRESH-Prompt} & \scriptsize{86.47±0.99} & \scriptsize{84.15±1.29} & \scriptsize{89.84±0.87} & \scriptsize{88.45±0.64} & \scriptsize{88.18±0.83} & \scriptsize{92.73±0.25}
        \\

        \hline

        & \multicolumn{6}{c}{Drug} \\
        \hline
        
        \scriptsize{Bio\_ClinicalBERT}  & \scriptsize{84.49±0.36} & \scriptsize{82.28±0.61} & \scriptsize{92.48±0.30} & \scriptsize{86.36±2.59} & \scriptsize{86.66±1.60} & \scriptsize{93.70±0.51}
        
        \\
        
        \scriptsize{Bio\_ClinicalBERT-KIRESH} & \scriptsize{85.52±1.81} & \scriptsize{84.15±3.70} & \scriptsize{93.33±0.76} &  \scriptsize{87.12±1.56} & \scriptsize{87.67±2.02} & \scriptsize{94.32±0.44}
        \\
        
        \scriptsize{Bio\_ClinicalBERT-Prompt} & \scriptsize{87.36±1.59} & \scriptsize{85.16±2.56} & \scriptsize{93.55±0.88} & \scriptsize{88.40±1.66} & \scriptsize{87.82±2.67} & \scriptsize{94.16±0.98}
        \\
        
        \scriptsize{Bio\_ClinicalBERT-KIRESH-Prompt} & \scriptsize{89.15±0.49} & \scriptsize{87.01±1.80} & \scriptsize{94.24±0.56} & \scriptsize{88.72±1.73} & \scriptsize{88.46±1.35} & \scriptsize{94.68±0.60}
        \\

        \hline
        
        \end{tabular}
        \caption{Experimental results of three SBDH in public MIMIC-SBDH dataset. The results show the generalizability of our proposed methods in SBDH prediction.}
        \label{Table:generalizability}
    \end{table*}

\end{document}